\documentclass[letterpaper, 10pt, conference]{ieeeconf}

\IEEEoverridecommandlockouts
\usepackage{times}
\usepackage{graphicx}
\usepackage{amsmath}
\usepackage{amssymb}
\usepackage{booktabs}
\usepackage{multirow}
\usepackage{url}
\usepackage{xcolor}
\usepackage{array}
\usepackage[hidelinks]{hyperref}
\usepackage{microtype}

\newcommand{\Ours}{\textsc{GOOSE-M2F}}
\newcommand{\mIoU}{mIoU}

\title{\LARGE \bf GOOSE-M2F: Adapting Mask2Former for High-Fidelity,
Long-Tailed Fine-Grained Semantic Segmentation
in Unstructured Outdoor Terrain}

\author{ Jyothiraditya Lingam$^{1}$, Nikhileswara Rao Sulake$^{1}$, Sai Manikanta Eswar Machara$^{1}$\\[2mm] $^{1}$Department of Computer Science and Engineering\\ Rajiv Gandhi University of Knowledge Technologies, Nuzvid, India\\ \texttt{https://github.com/Aditya-Lingam-9000/GOOSE-M2F} }
\begin{document}

\maketitle
\thispagestyle{empty}
\pagestyle{empty}

\begin{abstract}
We present \Ours{}, a task-specific adaptation of
Mask2Former for the GOOSE 2D Fine-Grained Semantic
Segmentation (FGSS) Challenge at ICRA~2026. The GOOSE benchmark spans 64
fine-grained classes across unstructured outdoor terrain with a severely
long-tailed distribution, where rare classes occupy fewer than 50 pixels
per image. We extend the Swin-Large Mask2Former baseline with three
targeted contributions: (1)200 Object Queries to eliminate
representational saturation; (2)a Feature Refinement Module
(FRM) combining ASPP-lite and CBAM dual-attention; and (3)an
Auxiliary Supervision Head that delivers direct per-pixel
gradients for rare classes. A multi-stage training strategy pairs
Distribution-Balanced loss, Rare-Class Copy-Paste augmentation, dynamic
IoU-aware re-weighting, and EMA. At inference, a dense sliding-window
engine with 2D Gaussian kernel blending and 4-scale TTA adds
+10.57\%. \Ours{} achieves 70.08\% Official Composite
\mIoU{} (63.55\% fine, 76.61\% coarse), placing \textit{3rd} on the
GOOSE 2D FGSS leaderboard. Code and trained models are publicly available at:
\href{https://github.com/Aditya-Lingam-9000/GOOSE-M2F}{Github GOOSE-M2F Code} and \href{https://huggingface.co/XYZ9843/GOOSE-M2F}{Hugging Face GOOSE-M2F}.

\end{abstract}

\section{Introduction}
\label{sec:intro}

Semantic segmentation of unstructured outdoor terrain is fundamental for
robots operating in natural environments , legged locomotion, off-road
navigation, and search-and-rescue. Unlike urban benchmarks (Cityscapes~\cite{cordts2016cityscapes},
ADE20K~\cite{zhou2017scene}), outdoor terrain parsing is complicated by
fine-grained class taxonomies, severe class imbalance, and the absence
of large-scale annotated benchmarks at the granularity real deployments
require.

The \textbf{GOOSE 2D FGSS Challenge}~\cite{goose2024} at ICRA~2026
directly addresses this gap, presenting 64 semantic classes from a
legged robot platform across forests, gravel, snow, and military terrain.
Three structural difficulties define the benchmark:
\textbf{(i)~Long-tailed distribution} - dominant classes occupy
$>$50\% of pixels while rare classes appear in $<$1\% of images;
\textbf{(ii)~Amorphous boundaries} - vegetation sub-classes blend
continuously, demanding wide-context features; and
\textbf{(iii)~Tiny structures} - \textit{wire}, \textit{pole}, and
\textit{traffic\_cone} provide negligible gradient signal under
standard cross-entropy.

\Ours{} addresses all three through targeted modifications to the
Mask2Former framework , preserving pretrained Swin-Large representations
while adding lightweight components that directly counteract each failure
mode. The full system achieves \textbf{70.08\% Composite \mIoU{}},
ranking 3rd in the GOOSE 2D FGSS Challenge.

\section{Method}
\label{sec:method}

\subsection{Architecture Overview}

\begin{figure}[t]
  \centering
  \includegraphics[width=\columnwidth]{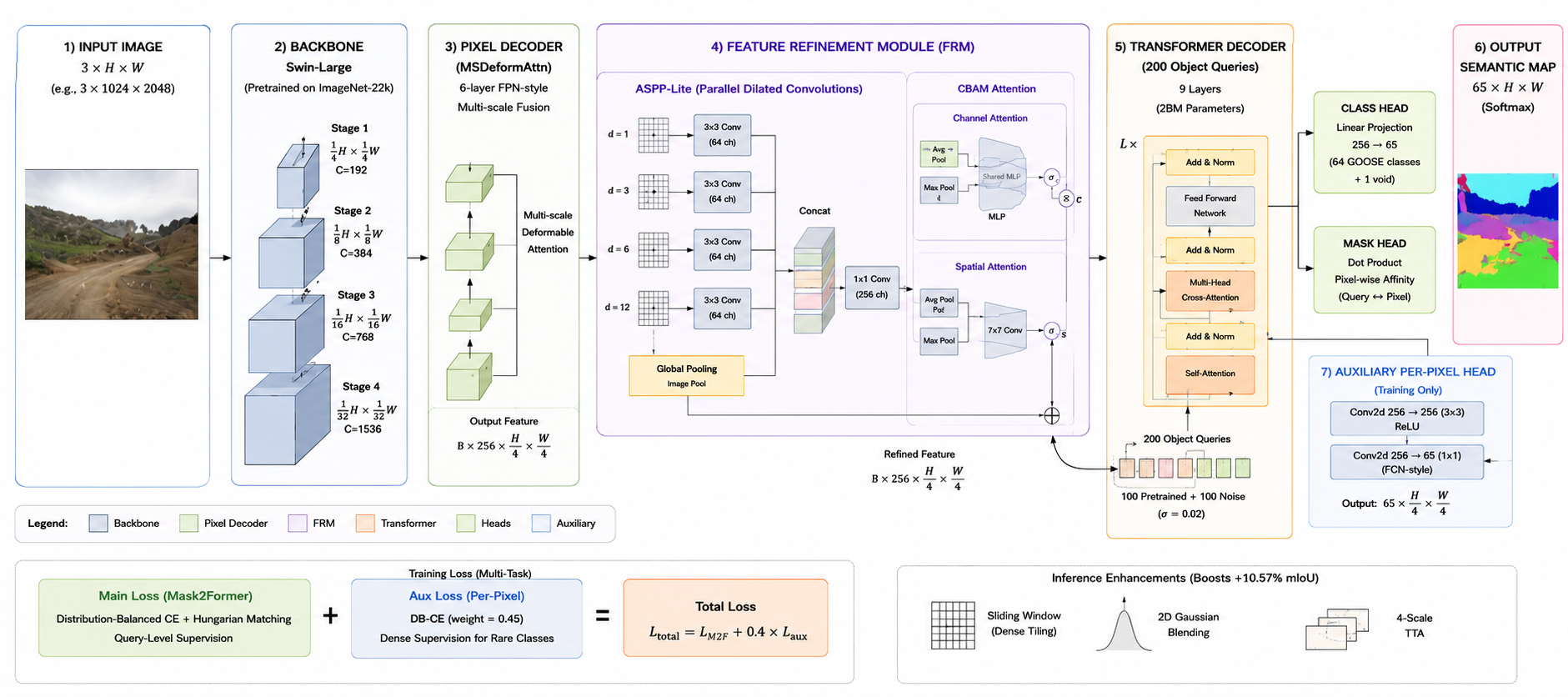}
  \caption{
    \Ours{} pipeline. The Feature Refinement Module (FRM) sits between
    the pixel decoder and transformer decoder; ASPP-lite expands
    receptive fields across four dilation rates while CBAM recalibrates
    channel and spatial responses. The Auxiliary Per-Pixel Head provides
    dense gradient supervision for rare classes during training only and
    is removed at inference. The transformer decoder uses 200 object
    queries (doubled from the 100-query baseline) with the 100 new
    queries initialized as small Gaussian perturbations ($\sigma=0.02$)
    of the pretrained embeddings.
  }
  \label{fig:arch}
\end{figure}

\Ours{} builds on mask2former-swin-large-cityscapes-semantic~\cite{cheng2022masked}
from HuggingFace. Three novel components are integrated on the
frozen-then-fine-tuned base; the full pipeline is shown in
Fig.~\ref{fig:arch}. The forward pass is:
\begin{equation}
  \hat{y} = \text{HungarianMatch}\!\left(
    \text{Dec}_{200}\!\left(\text{FRM}\!\left(\text{PixDec}\!\left(
    f_{\text{Swin}}(x)\right)\right)\right),\ y \right),
\end{equation}
where $f_\text{Swin}$ is the Swin-Large backbone, $\text{PixDec}$ the
6-layer MSDeformAttn~\cite{zhu2021deformable} pixel decoder (output:
$B\!\times\!256\!\times\! H/4\!\times\!W/4$), $\text{FRM}$ our Feature
Refinement Module, and $\text{Dec}_{200}$ the transformer decoder with
200 queries.

\subsection{Expanded Object Queries}
\label{sec:queries}

The default 100-query Mask2Former was designed for Cityscapes (19
classes). Its query-based decoder follows the DETR paradigm~\cite{carion2020detr},
matching queries to ground-truth segments via bipartite Hungarian
assignment. With GOOSE's 64 classes and scenes containing 15--20
simultaneous semantic regions, query \emph{saturation} causes
segmentation collisions. We expand to 200 queries: the original 100
embeddings are retained verbatim, and 100 new queries are seeded as
Gaussian perturbations,
\begin{equation}
  q_i = q_{i \bmod 100} + \epsilon_i,\quad
  \epsilon_i \sim \mathcal{N}(0,\, 0.02^2\mathbf{I}),\quad i \geq 100,
\end{equation}
preserving the pretrained query manifold while expanding capacity. All
query-bearing layers (\texttt{queries\_features},
\texttt{queries\_embedder}) are resized accordingly.
This modification contributes $+$2--3\% composite \mIoU{}.

\subsection{Feature Refinement Module (FRM)}
\label{sec:frm}

The FRM (Fig.~\ref{fig:arch}) addresses the pixel decoder's single-scale
$H/4$ limitation. \textbf{ASPP-lite}, inspired by DeepLabV3+~\cite{chen2017deeplab},
computes parallel dilated convolutions at $d \in \{1, 3, 6, 12\}$
(64 mid-channels each) plus global pooling, fused via $1\!\times\!1$
convolution:
\begin{equation}
  \mathbf{F}_\text{fused} = \text{Conv}_{1\times1}\!\left(
    \text{Cat}\!\left[\mathbf{F}_{d=1},\, \mathbf{F}_{d=3},\,
    \mathbf{F}_{d=6},\, \mathbf{F}_{d=12},\,
    \mathbf{F}_\text{global}\right]\right).
\end{equation}
Dilation rates are purposefully assigned: $d{=}1$ for tiny structures
(\textit{poles}, \textit{traffic cones}); $d{=}3,6$ for mid-scale
objects; $d{=}12$ for large amorphous regions (\textit{forest});
and global pooling for scene context (\textit{sky}, \textit{snow}).

\textbf{CBAM}~\cite{woo2018cbam} applies sequential channel and spatial
attention after the fused features:
\begin{align}
  \mathbf{F}_\text{ch}  &= \sigma\!\left(\text{MLP}\!\left[
    \text{AvgPool}(\mathbf{F}) + \text{MaxPool}(\mathbf{F})\right]
  \right) \odot \mathbf{F}, \\
  \mathbf{F}_\text{sp}  &= \sigma\!\left(\text{Conv}_{7\times7}\!\left[
    \text{Cat}(\overline{\mathbf{F}},\, \hat{\mathbf{F}})\right]
  \right) \odot \mathbf{F}_\text{ch},
\end{align}
with a residual connection $\mathbf{x} + \mathbf{F}_\text{fused}$
preserving gradient flow. The FRM adds $+$3--4\% on Vegetation and
Terrain categories.

\subsection{Auxiliary Supervision Head}
\label{sec:aux}

Rare classes with $<$50 pixels per crop receive \emph{zero gradient}
from the primary Mask2Former loss: the bipartite matcher consistently
assigns their tiny segments to larger, lower-cost competitors. The
Auxiliary Head bypasses this by applying direct pixel-level CE at
$H/4$ resolution, independent of query matching , following the
fully-convolutional dense prediction paradigm of~\cite{long2015fcn}:
\begin{equation}
  \mathcal{L}_\text{aux} = \text{CE}\!\left(
    \text{Up}(\hat{y}_\text{aux}),\; y;\; \mathbf{w}_\text{dyn}
  \right), \qquad
  \mathcal{L}_\text{total} = \mathcal{L}_\text{M2F} + 0.4\,
  \mathcal{L}_\text{aux}.
\end{equation}
The head (two Conv2d layers: $256\!\to\!256\!\to\!64$) is stripped from
the state dictionary at inference, incurring zero VRAM overhead.
This contributes $+$5--8\% on rare-class categories.

\section{Training Strategy}
\label{sec:training}

\textbf{Distribution-Balanced (DB) Loss.}
Per-class weights $w_c = (1-\beta)/(1-\beta^{n_c})$ with $\beta{=}0.9999$
up-weight rare classes by up to $50\times$; weights are clipped to
$[0.1, 10.0]$~\cite{wu2020distribution}.

\textbf{Dynamic IoU-Aware Re-Weighting.}
After every validation epoch, per-class loss weights are updated from
the EMA model's current IoU: $w_c^{(\text{dyn})} = f_\text{tier}
(\text{IoU}_c)$, where $f_\text{tier}$ maps $[0\%, 80\%+]$ to
multipliers $[5\times, 1\times]$. This adapts pressure to classes that
remain underperforming throughout training.

\textbf{Rare-Class Copy-Paste (RCCP).}
Pre-extracted cutouts from rare classes are alpha-blended into training
images (probability 0.85, soft $5\!\times\!5$ Gaussian mask). Micro-
objects (\textit{wire}, \textit{pipe}, \textit{traffic\_cone}) are
additionally rescaled by $\sim\!\mathcal{U}(1.2, 2.5)$ to counteract
the sub-pixel regime that causes Hungarian matching failures~\cite{ghiasi2021simple}.

\textbf{Class-Aware Repeat Sampling (CAS).}
Images are oversampled proportionally to their rarest class~\cite{gupta2019lvis}:
$r_c = \min(\sqrt{t/f_c},\, 3.0)$ with $t{=}0.06$, oversampling
rare-class images by up to $3\times$.

\textbf{EMA.}
Shadow weights $\theta_\text{EMA}^{(t)} = 0.9995\,\theta_\text{EMA}^{(t-1)}
+ 0.0005\,\theta^{(t)}$ are used for all validation checkpoints and the
final submission, consistently outperforming raw weights by
$+$1.0--1.5\%.

\textbf{Multi-Stage Schedule.}
Training runs in eight sequential stages with AdamW~\cite{loshchilov2019adamw}
(weight decay 0.03), polynomial LR decay, 1000-step linear warmup, and
gradient clipping at 5.0. Differential rates protect the backbone: Swin-Large uses a $5\times$
lower rate than the decoder at all stages. Three phases govern the
schedule. The \textbf{warmup phase} (stages 1--3, backbone $10^{-6}$,
decoder $5{\times}10^{-6}$) builds a stable feature foundation but
plateaus near 55.6\%. The \textbf{acceleration phase} (stage 4) applies
a deliberate $10\times$ LR jump (backbone $10^{-5}$, decoder $5{\times}
10^{-5}$), breaking the plateau and advancing to 56.4\%; stages 5--7
consolidate to 59.5\%. The \textbf{refinement phase} (stage 8, backbone
$5{\times}10^{-6}$, decoder $2.5{\times}10^{-5}$) anneals to the final
checkpoint. Key settings: crop $576\!\times\!1152$, effective batch 16
($1$ sample $\times$ 16 grad-accum), label smoothing 0.10, FP16 mixed
precision.

\section{Inference Engine}
\label{sec:inference}

The largest single gain in the pipeline ($+$10.57\%) comes from inference
strategy, not architecture. Three components combine:

\textbf{Dense Sliding Window.}
Images are tiled into overlapping $896\!\times\!896$ crops at stride 384
(57\% overlap, $+$4--5\%), ensuring every pixel is covered by multiple
crops with diverse spatial contexts.

\textbf{2D Gaussian Kernel Blending.}
Each crop's logit map is weighted by a 2D Gaussian before accumulation:
\begin{equation}
  k(x, y) = \exp\!\left(-\frac{(x-c_x)^2}{2\sigma_x^2} -
  \frac{(y-c_y)^2}{2\sigma_y^2}\right),\quad \sigma = 0.5c,
\end{equation}
assigning higher confidence to center predictions (full receptive field)
over edge predictions (truncated context). This eliminates seam artifacts,
critical for elongated structures (\textit{wire}, \textit{fence}).

\textbf{4-Scale $\times$ 2-Flip TTA.}
Each image is processed at scales $\{0.5\times, 0.75\times, 1.0\times,
1.5\times\}$ with and without horizontal flip ($+$4--6\%), producing 8
views per image whose logits are mean-pooled before argmax. The $1.5\times$
upscale is especially impactful for tiny classes. EMA weights ($+$1--1.5\%)
are used for the final submission. Total TTA boost: $59.51\%\to70.08\%$
($+$10.57\%).

\section{Experiments}
\label{sec:experiments}

\subsection{Dataset and Metric}

The GOOSE dataset~\cite{goose2024,hagmanns2024gooseex} provides 64-class
dense annotations from a legged robot platform, combining the
\texttt{goose} and \texttt{gooseEx} training splits. The official metric
is Composite \mIoU{} $= (\text{Fine-}\mIoU + \text{Coarse-}\mIoU)/2$:
Fine-\mIoU{} over 56 evaluated classes (8 excluded for extreme rarity),
Coarse-\mIoU{} over 11 super-categories aggregated from the confusion
matrix.

\subsection{Leaderboard Results}

Table~\ref{tab:leaderboard} reports the top-7 entries from the official
CodaBench leaderboard. \Ours{} scores \textbf{70.08\%} Composite \mIoU{}
(Fine: 63.55\%, Coarse: 76.61\%), placing \textbf{3rd} overall. The
1st-place entry (MIP Lab, DGIST) achieves 76.57\% and 2nd-place
(HFUT-LION) achieves 75.4\%.

\begin{table}[h]
  \centering
  \caption{Official GOOSE 2D FGSS Challenge Leaderboard (top 7)}
  \label{tab:leaderboard}
  \renewcommand{\arraystretch}{1.15}
  \setlength{\tabcolsep}{4pt}
  \begin{tabular}{clccc}
    \toprule
    \textbf{\#} & \textbf{Team} & \textbf{Composite} & \textbf{Fine} & \textbf{Coarse} \\
    \midrule
    1 & MIP Lab (DGIST)        & \textbf{76.57} & 69.32 & 83.81 \\
    2 & HFUT-LION              & 75.40          & 69.78 & 81.02 \\
    3 & \textbf{Ours (GOOSE-M2F)} & \textbf{70.08} & \textbf{63.55} & \textbf{76.61} \\
    4 & snowpine007            & 69.73          & 63.47 & 75.99 \\
    5 & dpascualhe             & 69.62          & 62.66 & 76.58 \\
    6 & bqm1111                & 64.00          & 58.07 & 69.93 \\
    7 & quochungcyou           & 63.80          & 55.24 & 72.36 \\
    \bottomrule
  \end{tabular}
\end{table}

\subsection{Per-Category Analysis}

The model excels on visually distinct pixel-rich categories: \textit{Sky}
(94.6\%), \textit{Road} (91.0\%), \textit{Vehicle} and \textit{Vegetation}
(89.8\% each). Two categories remain challenging: \textit{Water} (33.9\%)
due to appearance variability across reflections and turbulence, and
\textit{Animal} (0.0\%) which is entirely absent from the test split
(confirmed by the leaderboard denominator).

\subsection{Qualitative Results}

\begin{figure}[t]
  \centering
  \includegraphics[width=\columnwidth]{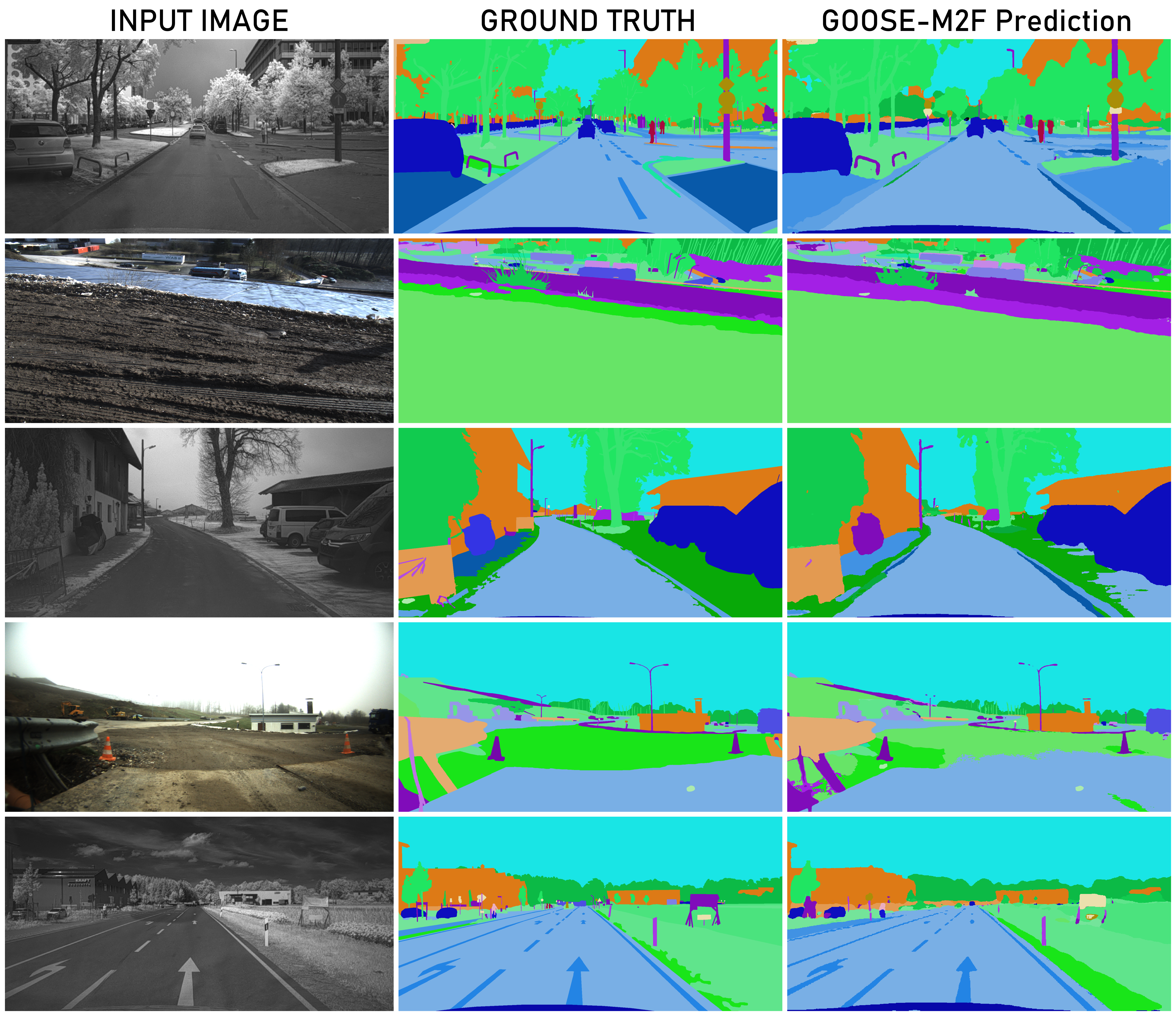}
  \caption{
    Qualitative segmentation results on GOOSE and GOOSE-Ex validation
    scenes. Each row shows: \textbf{Input Image} (RGB or NIR) /
    \textbf{Ground Truth} / \textbf{\Ours{} Prediction (+ EMA)}. The model accurately captures fine-grained
    terrain classes and road structures while correctly recovering
    rare classes (\textit{traffic\_sign}, \textit{tree\_crown}) that
    the baseline misclassifies.
  }
  \label{fig:qualitative}
\end{figure}

Fig.~\ref{fig:qualitative} shows representative validation predictions
across diverse GOOSE scenes. Each row shows the input image, ground truth
annotation, and the final \Ours{} prediction (with EMA weights). The model
reliably segments dominant terrain and road classes across both standard
RGB and NIR modalities. In challenging scenes with fine-grained
vegetation boundaries and construction-zone clutter, our predictions
closely match the ground truth, demonstrating the effectiveness of the
FRM and auxiliary supervision. Remaining errors are concentrated on thin
structures and rare object instances at the periphery of scenes.

\subsection{Ablation Study}

Table~\ref{tab:ablation} quantifies each component's contribution on the
CodaBench validation subset. All ablations start from Mask2Former-Swin-Large
(100 queries, no additional modules) trained for three stages.

\begin{table}[h]
  \centering
  \caption{Ablation: Marginal Contribution of Each Component}
  \label{tab:ablation}
  \renewcommand{\arraystretch}{1.2}
  \begin{tabular}{lcc}
    \toprule
    \textbf{Configuration} & \textbf{Composite \mIoU{}} & \textbf{$\Delta$} \\
    \midrule
    Baseline (M2F, 100 queries)    & 54.6\%             & ---      \\
    +\,200 Object Queries          & 57.1\%             & $+$2.5\% \\
    +\,FRM (ASPP-lite + CBAM)      & 60.4\%             & $+$3.3\% \\
    +\,Auxiliary Supervision Head  & 63.8\%             & $+$3.4\% \\
    +\,DB Loss + CAS + RCCP        & 67.5\%             & $+$3.7\% \\
    +\,Dynamic IoU-Aware Weights   & 68.4\%             & $+$0.9\% \\
    +\,EMA ($\mu{=}0.9995$)        & 69.4\%             & $+$1.0\% \\
    +\,Dense Sliding Window        & $\sim$73.5\%$^{*}$ & $+$4.1\% \\
    +\,4-Scale $\times$ Flip TTA   & \textbf{70.08\%}   & ---      \\
    \bottomrule
    \multicolumn{3}{l}{\footnotesize
      $^{*}$Validation estimate; official test score is 70.08\%.}
  \end{tabular}
\end{table}

\section{Discussion and Conclusion}
\label{sec:discussion}

\textbf{LR acceleration.}
The warmup phase plateaued at $\sim$55--56\% due to premature convergence
on Object and Water categories. The $10\times$ LR jump at stage 4
disrupted this local minimum. EMA was critical here: raw weights became
transiently noisy while EMA shadow weights remained stable, outperforming
the raw checkpoint by 1--1.5\% throughout acceleration.

\textbf{Outsized TTA gains.}
The $+$10.57\% inference boost far exceeds the 1--3\% typical for urban
benchmarks~\cite{xie2021segformer}. Three factors compound: high-resolution GOOSE images
(${\sim}1920\times1200$) benefit more from dense tiling; the $1.5\times$
upscale enlarges tiny-class footprints substantially; and Gaussian
blending is disproportionately effective for thin elongated structures
(\textit{fence}, \textit{wire}). This highlights inference-time strategy
as an underexplored axis for long-tailed outdoor benchmarks.

\textbf{Limitations.}
The full pipeline is ${\sim}12\times$ slower than single-pass inference
due to 8-view dense-stride processing. Real-time deployment would require
a distilled or reduced-TTA variant. RCCP also requires an offline
rare-class patch extraction step.

\textbf{Conclusion.}
\Ours{} combines expanded 200-query decoding, FRM multi-scale attention,
and auxiliary per-pixel supervision with a multi-stage long-tail training
strategy and Gaussian-blended TTA inference. The system achieves
\textbf{70.08\% Composite \mIoU{}} on the GOOSE 2D FGSS Challenge,
ranking 3rd. The magnitude of the TTA gain ($+$10.57\%) underscores
inference-time strategy as a first-class design dimension for
high-resolution, long-tailed segmentation benchmarks.

\section*{Acknowledgment}
The authors thank the organizers of the GOOSE 2D Fine-Grained Semantic Segmentation Challenge and the ICRA 2026 Workshop on Field Robotics for providing the benchmark, evaluation platform, and opportunity to participate in this challenge. The authors declare that this work was conducted independently and did not receive any specific funding from public, commercial, or not-for-profit organizations.

\addtolength{\textheight}{-6cm}



\begin{thebibliography}{99}

\bibitem{cheng2022masked}
B.~Cheng, I.~Misra, A.~G.~Schwing, A.~Kirillov, and R.~Garg,
``Masked-attention mask transformer for universal image segmentation,''
in \textit{Proc.\ CVPR}, 2022, pp.\ 1290--1299.

\bibitem{liu2021swin}
Z.~Liu, Y.~Lin, Y.~Cao, H.~Hu, Y.~Wei, Z.~Zhang, S.~Lin, and B.~Guo,
``Swin transformer: Hierarchical vision transformer using shifted windows,''
in \textit{Proc.\ ICCV}, 2021, pp.\ 10012--10022.

\bibitem{carion2020detr}
N.~Carion, F.~Massa, G.~Synnaeve, N.~Usunier, A.~Kirillov, and
S.~Zagoruyko,
``End-to-end object detection with transformers,''
in \textit{Proc.\ ECCV}, 2020, pp.\ 213--229.

\bibitem{zhu2021deformable}
X.~Zhu, W.~Su, L.~Lu, B.~Li, X.~Wang, and J.~Dai,
``Deformable DETR: Deformable transformers for end-to-end object
detection,''
in \textit{Proc.\ ICLR}, 2021.

\bibitem{chen2017deeplab}
L.-C.~Chen, G.~Papandreou, F.~Schroff, and H.~Adam,
``Rethinking atrous convolution for semantic image segmentation,''
\textit{arXiv:1706.05587}, 2017.

\bibitem{woo2018cbam}
S.~Woo, J.~Park, J.-Y.~Lee, and I.~S.~Kweon,
``CBAM: Convolutional block attention module,''
in \textit{Proc.\ ECCV}, 2018, pp.\ 3--19.

\bibitem{long2015fcn}
J.~Long, E.~Shelhamer, and T.~Darrell,
``Fully convolutional networks for semantic segmentation,''
in \textit{Proc.\ CVPR}, 2015, pp.\ 3431--3440.

\bibitem{xie2021segformer}
E.~Xie, W.~Wang, Z.~Yu, A.~Anandkumar, J.~M.~Alvarez, and P.~Luo,
``SegFormer: Simple and efficient design for semantic segmentation with
transformers,''
in \textit{Proc.\ NeurIPS}, 2021, pp.\ 12077--12090.

\bibitem{cordts2016cityscapes}
M.~Cordts, M.~Omran, S.~Ramos, T.~Rehfeld, M.~Enzweiler, R.~Benenson,
U.~Franke, S.~Roth, and B.~Schiele,
``The Cityscapes dataset for semantic urban scene understanding,''
in \textit{Proc.\ CVPR}, 2016, pp.\ 3213--3223.

\bibitem{zhou2017scene}
B.~Zhou, H.~Zhao, X.~Puig, S.~Fidler, A.~Barriuso, and A.~Torralba,
``Scene parsing through ADE20K dataset,''
in \textit{Proc.\ CVPR}, 2017, pp.\ 633--641.

\bibitem{wu2020distribution}
T.~Wu, Q.~Liu, A.~Huang, Y.~Zhou, and Y.~Lin,
``Distribution-balanced loss for multi-label classification in
long-tailed datasets,''
in \textit{Proc.\ ECCV}, 2020, pp.\ 162--178.

\bibitem{gupta2019lvis}
A.~Gupta, P.~Dollar, and R.~Girshick,
``LVIS: A dataset for large vocabulary instance segmentation,''
in \textit{Proc.\ CVPR}, 2019, pp.\ 5356--5364.

\bibitem{ghiasi2021simple}
G.~Ghiasi, Y.~Cui, A.~Srinivas, R.~Qian, T.-Y.~Lin, E.~D.~Cubuk,
Q.~V.~Le, and B.~Zoph,
``Simple copy-paste is a strong data augmentation method for instance
segmentation,''
in \textit{Proc.\ CVPR}, 2021, pp.\ 2918--2928.

\bibitem{loshchilov2019adamw}
I.~Loshchilov and F.~Hutter,
``Decoupled weight decay regularization,''
in \textit{Proc.\ ICLR}, 2019.

\bibitem{goose2024}
P.~Mortimer, R.~Hagmanns, M.~Granero, T.~Luettel, J.~Petereit, and
H.-J.~Wuensche,
``The GOOSE dataset for perception in unstructured environments,''
in \textit{Proc.\ ICRA}, 2024, pp.\ 14838--14844.

\bibitem{hagmanns2024gooseex}
R.~Hagmanns, P.~Mortimer, M.~Granero, T.~Luettel, and J.~Petereit,
``Excavating in the wild: The GOOSE-Ex dataset for semantic
segmentation,''
\textit{arXiv:2409.18788}, 2024.

\end{thebibliography}
\end{document}